\documentclass[letterpaper, 10 pt, conference]{ieeeconf}  

\IEEEoverridecommandlockouts                              

\usepackage[table,xcdraw,dvipsnames]{xcolor}

\usepackage{cite}
\usepackage{graphicx}
\usepackage{epsfig} 
\usepackage{mathptmx}
\usepackage{amsmath} 
\usepackage{wrapfig}
\usepackage{capt-of}

\usepackage{subcaption}

\usepackage{amssymb}  
\usepackage{amsthm}
\usepackage{wasysym}
\usepackage{mathrsfs}
\usepackage{cancel}
\usepackage[mathcal]{euscript}
\usepackage{bbm}
\usepackage{color}
\usepackage{lipsum}
\usepackage{algorithm}
\usepackage{algpseudocode}
\usepackage{xspace}
\usepackage{float}
\usepackage{placeins}
\usepackage{makecell}
\usepackage{graphicx}
\usepackage[colorlinks=true,linkcolor=black,anchorcolor=black,citecolor=black,filecolor=black,menucolor=black,runcolor=black,urlcolor=epcc_blue]{hyperref}
\usepackage[normalem]{ulem}
\usepackage{tabularx}
\usepackage{booktabs}
\usepackage{array}
\usepackage{etoolbox}
\usepackage{multirow}

\usepackage{subcaption}
\usepackage{graphicx}

\usepackage{xfrac}

\newcolumntype{Y}{>{\centering\arraybackslash}X}
\DeclareMathOperator*{\argmin}{arg\,min}

\usepackage{outlines}
\let\labelindent\relax
\usepackage{enumitem}
\setenumerate[2]{label=\alph*.}
\setenumerate[3]{label=\roman*.}

\makeatletter
\patchcmd{\@makecaption}
  {\scshape}
  {}
  {}
  {}
\makeatother

\usepackage[capitalise]{cleveref}
\crefname{equation}{Eq.}{Eqs.}
\crefname{figure}{Fig.}{Figs.}
\crefname{tabular}{Tab.}{Tabs.}
\crefname{section}{Sec.}{Secs.}
\crefname{algorithm}{Alg.}{Algs.}
\crefname{appendix}{Appx.}{Appxs.}
\crefname{thm}{Thm.}{Thms.}
\crefname{cor}{Cor.}{Cors.}
\crefname{algorithm}{Alg.}{Algs.}
\crefname{lemma}{Lem.}{Lems.}
\crefname{problem}{Prob.}{Probs.}

\newcommand{\regtext}[1]{\mathrm{\textnormal{#1}}}

\newcommand{\lbl}[1]{_{\regtext{#1}}}

\newcommand{\vc}[1]{#1}
\newcommand{\set}[1]{\mathcal{#1}}

\newcommand{\expectation}{\mathbb{E}}

\newcommand{\dataset}{\set{D}}
\newcommand{\trajectory}{\tau}

\newcommand{\expert}{\lbl{e}}
\newcommand{\play}{\lbl{p}}

\newcommand{\timestep}{t}
\newcommand{\horizon}{T}

\newcommand{\actionhorizon}{{\horizon_{\action}}}

\newcommand{\action}{\vc{a}}

\newcommand{\observation}{\vc{o}}
\newcommand{\latent}{\vc{z}}

\newcommand{\actionsequence}{\vc{A}}
\newcommand{\latentsequence}{\vc{Z}}

\newcommand{\actionspace}{\set{A}}

\newcommand{\observationspace}{\set{O}}
\newcommand{\latentspace}{\set{Z}}

\newcommand{\dynamics}{\vc{f}_{\theta}}

\newcommand{\policy}{\vc{\pi}}

\newcommand{\encoder}{h_\phi}
\newcommand{\ltg}{G_\psi}
\newcommand{\lcg}{C_\psi}

\definecolor{epcc_blue}{RGB}{0,114,178}
\newcommand{\ourmethod}{\textcolor{epcc_blue}{\textbf{\textsc{EPCC}}\xspace}}

\newcommand{\idx}[1]{^{(#1)}} 

\newcommand{\nocontentsline}[3]{}
\newcommand{\tocless}[2]{\bgroup\let\addcontentsline=\nocontentsline#1{#2}\egroup}

\newcommand{\DP}{\textsc{DP}\xspace}

\newcommand{\SAIL}{\textsc{SAIL}\xspace}

\newcommand{\WMFull}{\textsc{Play+Goal}\xspace}

\newcommand{\Ours}{\textcolor{epcc_blue}{\textbf{\textsc{EPCC}}}\xspace}
\newcommand{\SuP}{\textsc{SuP}\xspace}
\newcommand{\SKIP}{\textsc{SkiP}\xspace}

\newcommand{\PushT}{PushT\xspace}
\newcommand{\Cube}{Cube\xspace}

\newcommand{\LowFricPushT}{Slippery-PushT\xspace}
\newcommand{\TargetTask}{interaction-sensitive\xspace}
\usepackage[table]{xcolor}
\definecolor{tphl}{HTML}{EAF2FB}          

\title{\LARGE \bf
Expert-Play Contouring Control: \\
Faster-than-Demonstration Planning from Slow Expert and Fast Play
}

\author{
Anonymous Authors
}

\IEEEaftertitletext{%
  \begin{minipage}{\textwidth}
    \centering
    \includegraphics[width=\textwidth]
      {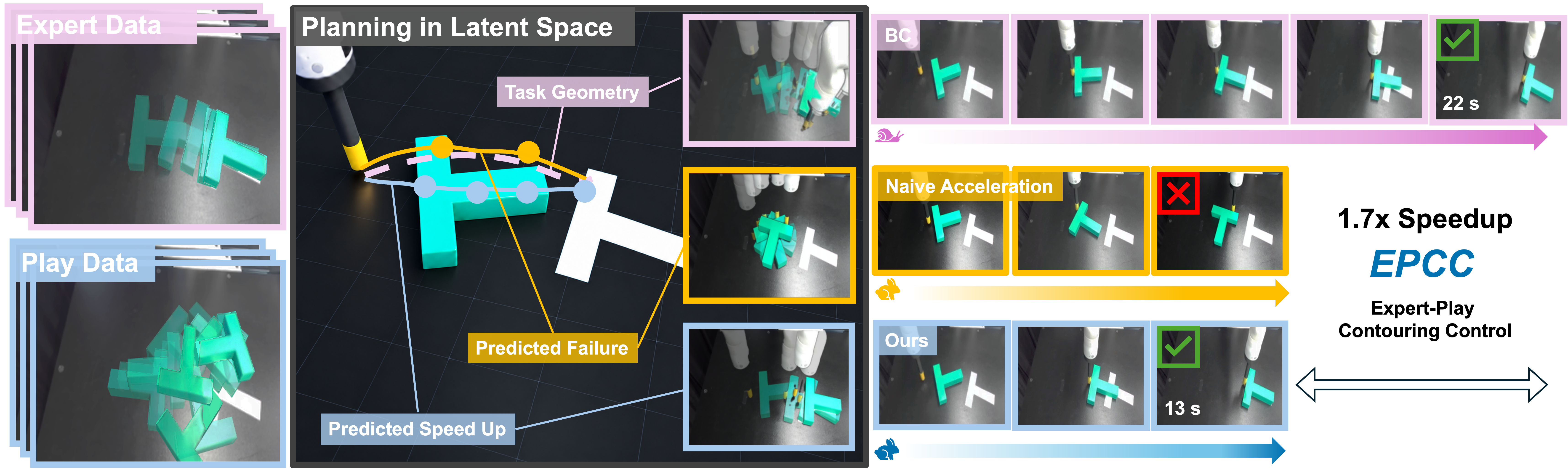}
    \captionof{figure}{\textbf{Expert-Play Contouring Control (\ourmethod)}: Expert demonstrations define the latent task geometry, while play data trains dynamics for predicting the outcomes of faster actions. 
    In latent space, EPCC evaluates candidate fast rollouts against the expert-defined task geometry, rejecting predicted failures and selecting actions that make faster task-consistent progress.
    As illustrated on the right, behavior cloning succeeds but remains slow, naive acceleration fails, and EPCC completes the task successfully with a $1.7\times$ speedup.
        }
    \label{fig:teaser}
    \vspace{0.5\baselineskip}
  \end{minipage}%
}
\author{%
Seunghoon Cho*$^{2}$,
Wonsuhk Jung*$^{1}$,
Sundhar Vinodh Sangeetha$^{1}$,
and Shreyas Kousik$^{1}$%
\thanks{$^{*}$These authors contributed equally.}%
\thanks{$^{1}$Wonsuhk Jung, Sundhar Vinodh Sangeetha,
and Shreyas Kousik are with the Georgia Institute
of Technology, Atlanta, GA, USA.
\texttt{wonsuhk.jung@gatech.edu},
\texttt{ssangeetha3@gatech.edu},
\texttt{shreyas.kousik@me.gatech.edu}.}%
\thanks{$^{2}$Seunghoon Cho is with Seoul National
University, Seoul, South Korea.
\texttt{csh020219@snu.ac.kr}.}%
}

\begin{document}

\maketitle

\thispagestyle{plain}
\pagestyle{plain}


\begin{abstract}
Expert demonstrations often specify what a robot should do, but not how fast it can do it. 
Imitation Learning (IL) inherits demonstration timing, while directly accelerating the learned motion can fail when faster execution changes the robot-object dynamics. 
We study faster-than-demonstration execution as a dynamics-aware control problem and introduce Expert-Play Contouring Control ($\ourmethod$), which combines slow expert demonstrations with fast, non-expert play.
Expert demonstrations train a latent trajectory generator whose predictions are reparameterized into a time-independent contour of successful task progression, while play trains a world model (WM) of fast-action outcomes.
At deployment, our planner uses the WM to optimize actions that maximize progress along the expert-derived contour while penalizing deviation from the intended task evolution.
Averaged across three visuomotor manipulation tasks, \Ours achieves $2.0\times$ the throughput of the IL baseline, including $2.3\times$ the throughput of the strongest acceleration baseline on a task with interaction-sensitive object dynamics.
Our analysis shows that the gains concentrate where faster execution changes robot-object evolution.
\Ours benefits from more play data, achieving higher success rates and throughput.
Together, our results highlight a simple yet effective principle: demonstrations provide task intent, while play data provides the dynamic coverage needed to execute that intent faster.
More details: \href{https://expert-play-contouring.netlify.app/}{https://expert-play-contouring.netlify.app/}.
\end{abstract}

\section{Introduction}

Imitation Learning (IL) provides a scalable framework for acquiring complex manipulation skills by avoiding hand-designing task objectives or system dynamics
\cite{moritz2023goalcondil, chi2025diffusion, barreiros2026careful}.
However, policies trained from human demonstrations often inherit their slow and cautious execution, limiting throughput \cite{bai2025towards}.
This limitation has motivated the study of \textit{faster-than-demonstration execution}
\cite{arachchige2025sail, guo2025demospeedup}.
A key challenge is that faster execution can induce robot-object interactions that are absent from slow demonstrations, pushing closed-loop rollouts outside the demonstrations' support.
Na\"ively executing an IL policy's predicted motion faster can therefore degrade task performance: the same motion need not induce the same task evolution at higher speed \cite{kim2026espada, kimtime}.
For example, a careful push that stops an object at its target goal on a low-friction surface may instead cause it to overshoot the target when executed faster (\Cref{fig:teaser}).

Because slow demonstrations provide no direct evidence of fast-action outcomes, most prior work mitigates the failure by identifying phases during which the motion generated by the policy can be executed faster without changing the outcome.
They use proxy cues for precision, such as motion complexity or semantic phase, and temporally compress the identified segments through action downsampling, faster tracking or retiming \cite{arachchige2025sail, guo2025demospeedup, dai2026skip, kim2026espada, kimtime}.
This strategy assumes that low-precision segments remain task-valid when the same underlying motion is executed faster.
However, proxy-based schedulers do not directly predict the resulting robot-object evolution, making them either conservative or unreliable and thereby limiting throughput \cite{wu2026speedup}.
More fundamentally, they can miss speedup opportunities in \TargetTask phases, where faster execution invalidates the original motion and preserving success requires adapting the motion itself.

In this paper, we investigate whether cheap non-expert play can provide the missing fast-regime action-outcome evidence, without requiring hand-designed analytic dynamics, thus preserving the scalability of IL.
Our key insight is to separate \textit{task intent} from \textit{fast-execution dynamics}: slow expert demos define the successful task evolution to preserve, whereas fast play reveals how actions change the robot-object evolution at higher speed (\Cref{fig:teaser}).
We integrate these complementary sources through a single planning objective that maximizes progress along the demonstrated path while keeping predicted rollouts close to it.
Instead of prescribing when and how much to accelerate, our method advances as far as the learned dynamics predict is feasible, enabling speedup during contact-sensitive interactions.

We call our approach \textbf{Expert-Play Contouring Control (\ourmethod)}.
Given the current observation, a latent contour generator trained from expert demonstrations produces a time-independent path of successful task progression, while a world model \cite{zhou2024dino, maes2026leworldmodel} learned from fast play predicts how candidate actions change the robot-object system.
Our Latent Contouring Controller uses these predictions to plan actions that maximize progress along the expert-derived contour.
By replanning online, \Ours accelerates whenever faster execution remains consistent with the intended task evolution and adapts when it does not.

We evaluate \Ours against an IL policy and three faster-than-demonstration baselines on three visuomotor manipulation tasks.
Averaged across tasks, \Ours achieves $2.0\times$ the throughput of the IL policy, the largest relative improvement among all evaluated methods.
Its advantage is greatest when faster actions substantially alter robot-object interactions: on an \TargetTask task, it achieves $2.3\times$ the throughput of the strongest acceleration baseline.
Across the three tasks, its contact-phase speedup is, on average, $1.15\times$ that of the strongest baseline.
Our ablations suggest that these gains arise from using play-trained dynamics to predict fast-action outcomes and optimizing actions beyond temporal rescaling of IL policy outputs.
We further characterize how the speed and amount of play data affect acceleration, finding that more data improves reliability.

To summarize, our contributions are:

\begin{enumerate}
    \item 
    We introduce Expert-Play Contouring Control, which combines an expert-derived latent task contour generator with a visual world model learned from non-expert play to optimize fast, task-progressing actions.

    \item
    Across three tasks, \Ours achieves the largest macro-averaged throughput gain over \DP ($2.0\times$), including $2.3\times$ the throughput of the strongest baseline on the \TargetTask \LowFricPushT task.

    \item We characterize when interaction-aware acceleration is beneficial, when synthesizing actions beyond temporally transforming policy outputs is important, and how the distribution and amount of play data affect performance.
\end{enumerate}

\section{Related Work}
\label{sec:related_works}

\textbf{Faster than Demonstration Execution.} 
Recent work has established faster-than-demonstration execution as an objective for imitation learning
\cite{arachchige2025sail,guo2025demospeedup}.
Most approaches accelerate motion specified by demonstrations or a base policy through temporal transformations, including faster tracking, demonstration or action-chunk downsampling, target skipping, and time-optimal retiming
\cite{arachchige2025sail,guo2025demospeedup,kim2026espada, dai2026skip,jing2026tempovla,yang2026realtime,kimtime}.
They typically use proxy cues such as motion complexity, action uncertainty, semantic phase, or robot-side feasibility to decide when and how aggressively these transformations should be applied.
Because these cues only indirectly indicate whether faster execution will preserve task progress, incorrect estimates can either miss feasible speedups or accelerate segments that no longer remain task-valid.

A complementary direction makes acceleration outcome-informed through task interaction or learned dynamics.
SpeedTuning learns acceleration through task-specific online RL, but still temporally transforms action chunks from a fixed base policy \cite{yuan2025speedtuning}.
\SuP is closest to our work: it uses a robot-state dynamics model to evaluate candidate downsampling rates, but remains confined to transformed chunks from a frozen policy and does not model visual object evolution
\cite{wu2026speedup}.
\Ours instead learns visual robot-object dynamics from offline non-expert play and uses them to optimize new control sequences beyond temporal transformations of demonstration or policy-derived motion.

\textbf{Planning with World Models.}
World Models (WMs) enable planning from visual observations by predicting action-conditioned future states in a compact latent space.
Most approaches plan through forward rollouts, optimizing actions to reach a terminal goal \cite{zhou2024dino, maes2026leworldmodel} or maximize a learned reward or value \cite{hansen2024td, hafner2023mastering}.
Concurrent work, LeFlow, instead amortizes planning by generating a goal-conditioned trajectory directly in the world-model latent space and decodes it into actions through inverse dynamics \cite{huang2026leflow}.
Our approach instead combines an expert-derived latent reference with forward model-based planning; expert demonstrations provide a latent contour of successful task progression, while forward world-model predictions determine which actions can advance along it and how quickly.

\textbf{Time-optimal Path Following Control.}
Model Predictive Contouring Control (MPCC) jointly optimizes control inputs and progress along a geometric contour, rewarding fast advancement while limiting deviation from the path
\cite{lam2010model,krinner_mpcc_2024}.
These methods, largely used in autonomous racing, assume that the relevant state, path, and dynamics are explicitly available, and primarily optimize the traversal of a prescribed robot path \cite{kimtime}.
In contrast, \Ours learns a task-level latent contour from expert demonstrations and visual robot-object dynamics from fast play, and uses them to jointly optimize actions and task progress in latent space.
\section{Problem Formulation}
\label{sec:problem}

We consider a visuomotor manipulation system with observation space
$\observationspace$ and action space $\actionspace$.
The robot is given two offline datasets.
The first is a set of successful expert demonstrations,
\begin{equation}
    \dataset\expert
    =
    \left\{
        \trajectory\expert\idx{i}
    \right\}_{i=1}^{N\expert},
    \qquad
    \trajectory\expert\idx{i}
    =
    \left\{
        \left(
            \observation_t\idx{i},
            \action_t\idx{i},
            \observation_{t+1}\idx{i}
        \right)
    \right\}_{t=0}^{\horizon\expert\idx{i}-1}.
    \label{eq:expert_dataset}
\end{equation}
Each expert trajectory completes the task, but need not execute it at
the fastest speed attainable by the robot.

The second dataset consists of play interactions,
\begin{equation}
    \dataset\play
    =
    \left\{
        \trajectory\play\idx{j}
    \right\}_{j=1}^{N\play},
    \qquad
    \trajectory\play\idx{j}
    =
    \left\{
        \left(
            \observation_t\idx{j},
            \action_t\idx{j},
            \observation_{t+1}\idx{j}
        \right)
    \right\}_{t=0}^{\horizon\play\idx{j}-1}.
    \label{eq:play_dataset}
\end{equation}
Play need not complete the task or exhibit expert
behavior; it provides experience of robot-object interactions under
actions that cover the target fast-execution regime.
We do not assume access to successful fast expert demonstrations.

Our objective is to synthesize a policy that rewards task success and short task completion time:

\begin{equation}
\begin{aligned}
    \policy^\star
    \in
    \argmin_{\policy}
    \quad
    &c_1\,
    \expectation
    \left[
        T(\trajectory_{\policy})
        \mid
        \mathsf S(\trajectory_{\policy})=1
    \right]
    \\
    &-
    c_2\,
    \expectation
    \left[
        \mathsf S(\trajectory_{\policy})
    \right],
    \qquad c_1,c_2>0 .
\end{aligned}
\label{eq:faster_than_demo_problem}
\end{equation}
Here $\trajectory_{\policy}$ is a closed-loop rollout,
$\mathsf S(\trajectory_{\policy})\in\{0,1\}$ indicates task success, and $T(\trajectory_{\policy})$ is completion time for a successful rollout.
The two terms respectively favor shorter successful rollouts and higher success rates, with $c_1$ and $c_2$ controlling the success-speed trade-off.
A target policy is \emph{faster than demonstration} if it completes successful rollouts faster than an imitation policy trained on expert demonstrations, while maintaining a comparable success rate.

The central challenge is that the two datasets provide different pieces of this objective.
Expert data identifies task evolutions that lead to success, whereas play data identifies how the robot-object system responds to actions in the fast regime.
Our method composes these complementary sources to infer fast actions without requiring fast successful demos.

\section{Method}
\label{sec:method}

\begin{figure*}[!t]
    \centering
    \includegraphics[width=\textwidth]{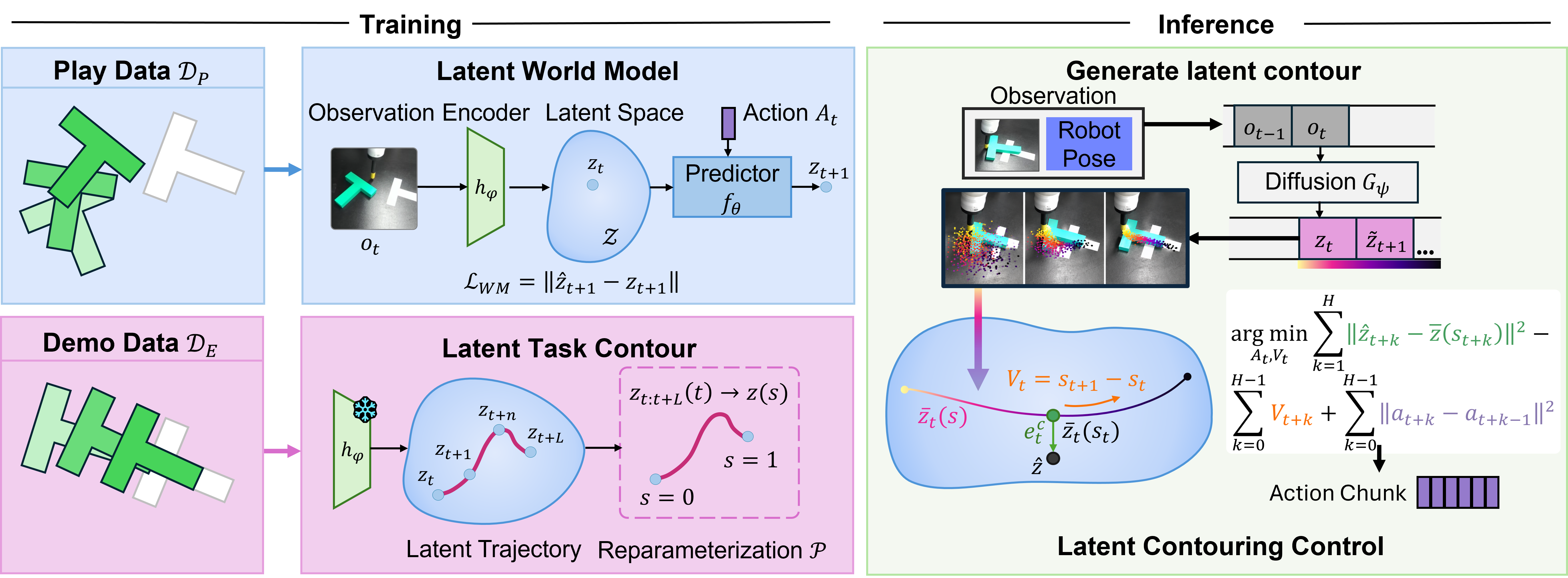}
    \vspace{-10pt}
    \caption{
    \textbf{Overview of \ourmethod.} \textbf{Training (left):} play data $\dataset\play$ trains a latent world model, consisting of an encoder $\encoder$ and an action-conditioned predictor $\dynamics$, with the next-latent prediction objective $\set{L}\lbl{WM}$ (\Cref{sec:world_model}). Expert demonstrations $\dataset\expert$ are encoded with the frozen $\encoder$ and used to train a latent trajectory generator $\ltg$; the path reparameterization $\set{P}$ removes the demonstrations' timing, yielding a time-independent contour $\bar{\latent}_\timestep(s)$, $s \in [0,1]$ (\Cref{sec:latent_contour}). \textbf{Inference (right):} from the current observation, the contour generator $\lcg$ produces the reference contour, and the controller jointly optimizes the action sequence $\actionsequence_\timestep$ and progress increments $V_\timestep$ by minimizing the contouring error $e^c_{\timestep+k|\timestep}$ between the world-model rollout $\hat{\latent}_{\timestep+k|\timestep}$ and $\bar{\latent}_\timestep(s_{\timestep+k|\timestep})$, maximizing progress $\Delta s$ along the contour, and regularizing the actions (\Cref{sec:epcc}, \Cref{eq:epcc_optimization}). The first $\actionhorizon$ actions are executed and both the contour and the plan are recomputed from the next observation.}
    \label{fig:method_overview}
    \vspace{-10pt}
\end{figure*}

In this section, we introduce \Ours to solve \eqref{eq:faster_than_demo_problem}.
We first learn a latent world model from fast play to predict robot-object dynamics, and a contour generator from expert demonstrations to produce an expert-derived latent contour from the current observation.
We then combine them through receding-horizon contouring control, which optimizes actions to maximize progress along the latent contour while keeping predicted rollouts close to it.
Please see \Cref{fig:method_overview} for overview.

\subsection{Learning Latent World Model from Play Data}
\label{sec:world_model}

Following action-conditioned latent world models
\cite{zhou2024dino, maes2026leworldmodel},
we encode visual observations into a compact latent state using $\encoder:\observationspace \rightarrow \mathcal Z$,
\begin{equation}
    \latent_t = \encoder(\observation_t),
    \label{eq:wm_encoder}
\end{equation}
and learn a latent dynamics model
$\dynamics:\latentspace \times \actionspace \rightarrow \latentspace$,
\begin{equation}
    \hat \latent_{t+1}
    =
    \dynamics(\latent_t,\action_t).
    \label{eq:wm_one_step}
\end{equation}

The world model is trained on play transitions using a next-latent prediction objective,
\begin{equation}
    \mathcal L_{\mathrm{WM}}(\phi,\theta)
    =
    \mathbb E_{(\observation_t,\action_t,\observation_{t+1})\sim \dataset\play}
    \left\|
        \dynamics\!\left(\encoder(\observation_t),\action_t\right)
        -
        \encoder(\observation_{t+1})
    \right\|_2^2
    \label{eq:wm_total_loss}
\end{equation}
The encoder may be fixed or learned jointly with the dynamics, with representation-specific regularization when needed.

At deployment time, an action sequence
\begin{equation}
    \actionsequence_{t:t+H-1}
    =
    \left(
        \action_{t\mid t},
        \action_{t+1\mid t},
        \ldots,
        \action_{t+H-1\mid t}
    \right)
\end{equation}
induces the predicted latent rollout
\begin{align}
    \hat \latent_{t\mid t}
    &=
    \encoder(\observation_t),
    \label{eq:wm_rollout_initial}
    \\
    \hat \latent_{t+k+1\mid t}
    &=
    \dynamics
    \left(
        \hat \latent_{t+k\mid t},
        \action_{t+k\mid t}
    \right),
    \qquad
    k=0,\ldots,H-1.
    \label{eq:wm_rollout_dynamics}
\end{align}
Because the encoder observes both the robot and the manipulated scene, the learned dynamics capture task-relevant robot-object evolution.

\subsection{Latent Contour Generation from Expert Demonstration}
\label{sec:latent_contour}
We construct a latent contour generator $\mathcal \lcg$ that maps the current latent state to a time-independent reference path in latent space.
The generator consists of two components: a learned latent trajectory generator $\ltg$ and a path reparameterization.
The former predicts the sequence of latent states that should be reached, while the latter removes their timings.

\textbf{Learning a latent trajectory generator.}
We freeze the world-model encoder $\encoder$ and use it to encode expert
demonstrations into the same latent space used by WM.
For an expert trajectory segment of length $L+1$, we define
\begin{equation}
    \latentsequence_{t:t+L}
    =
    \left(
        \latent_t,
        \latent_{t+1},
        \ldots,
        \latent_{t+L}
    \right),
    \qquad
    \latent_{t+k}=\encoder(\observation_{t+k}).
    \label{eq:expert_latent_trajectory}
\end{equation}
We then train a latent trajectory generator $\ltg$ to predict the future latent sequence from the current latent state:
\begin{equation}
    \mathcal L_{\mathrm{LTG}}(\psi)
    =
    \mathbb E_{\latentsequence_{t:t+L}\sim\dataset\expert}
    \left[
        \mathcal L_{\mathrm{traj}}
        \left(
            \ltg(\latent_t),
            \latentsequence_{t+1:t+L}
        \right)
    \right],
    \label{eq:latent_trajectory_generator_loss}
\end{equation}
where $\mathcal L_{\mathrm{traj}}$ denotes the training objective of the trajectory generator (e.g., diffusion, flow).

At inference, the current observation is encoded as $\latent_t=\encoder(\observation_t)$, and the generator predicts
\begin{equation}
    \ltg(\latent_t)
    =
    \left(
        \tilde \latent_{t+1\mid t},
        \ldots,
        \tilde \latent_{t+L\mid t}
    \right).
    \label{eq:generated_latent_future}
\end{equation}
Anchoring the prediction at the current latent state gives the generated latent trajectory
\begin{equation}
    \tilde \latentsequence_{t:t+L\mid t}
    =
    \left(
        \latent_t,
        \tilde \latent_{t+1\mid t},
        \ldots,
        \tilde \latent_{t+L\mid t}
    \right).
    \label{eq:generated_latent_trajectory}
\end{equation}
Here, $L$ denotes the reference horizon.
We choose $L$ to be sufficiently long to expose enough future task progression for meaningful acceleration.

\textbf{Path reparameterization.}
The generated latent trajectory $\tilde \latentsequence_{t:t+L\mid t}$ captures the desired future task evolution, but remains time-indexed and therefore inherits the timing of the expert demonstrations.
We convert it into a time-independent latent contour $\bar \latent_t:[0,1]\rightarrow\mathcal Z$ that preserves the ordering of the intermediate states while removing timing.

Using the latent distance $d_{\latentspace}(\latent,\latent'):=\|\latent-\latent'\|_2$, we assign each generated state a normalized cumulative path coordinate
\begin{align}
    s_{t,k}
    &:=
    \frac{\sum_{j=1}^{k}d_{\mathcal Z}
    \left(
        \tilde \latent_{t+j\mid t},
        \tilde \latent_{t+j-1\mid t}
    \right)}
         {\sum_{j=1}^{L}d_{\mathcal Z}
    \left(
        \tilde \latent_{t+j\mid t},
        \tilde \latent_{t+j-1\mid t}
    \right)},
    && k=1,\ldots,L,
    \label{eq:latent_path_coordinate}
\end{align}
with $\tilde \latent_{t\mid t}=\latent_t$ and $s_{t,0} = 0$.
For any continuous progress $s \in [s_{t,k}, s_{t, k+1}]$, we define the contour by piecewise-linear interpolation, exploiting latent straightening property \cite{maes2026leworldmodel},
\begin{equation}
    \bar \latent_t(s) 
    = 
    \bigl(1-\alpha\bigr)\tilde \latent_{t+k\mid t} 
    + 
    \alpha \tilde \latent_{t+k+1\mid t}, 
    \quad 
    \alpha =\frac{s-s_{t,k}}{s_{t,k+1}-s_{t,k}}
    \label{eq:latent_contour_interpolation}
\end{equation}
Thus, $\bar\latent_t(s_{t,k}) = \tilde \latent_{t+k|t}$, with $\bar\latent_t(0)=\latent_t$ and $\bar\latent_t(1)=\tilde\latent_{t+L \mid t}$.
The resulting contour preserves the geometry and ordering of the generated task evolution while leaving its traversal speed to the controller.
We denote this deterministic path reparameterization by
\begin{equation}
    \bar \latent_t(\cdot) = \mathcal P \left (\tilde \latentsequence_{t:t+L\mid t} \right )
    \label{eq:path_reparameterization_operator}
\end{equation}

Together with the learned trajectory generator, this defines the latent contour generator
\begin{equation}
    \mathcal \lcg(\latent_t)
    :=
    \mathcal P
    \left(
        \latent_t,
        \ltg(\latent_t)
    \right)
    =
    \bar \latent_t(\cdot),
    \label{eq:latent_contour_generator}
\end{equation}
which produces the time-independent reference contour used by \Ours for progress-aware control.

\subsection{Latent Contouring Control}
\label{sec:epcc}

Given the current latent state $\latent_t=\encoder(\observation_t)$, the latent contour generator produces a time-independent reference
\begin{equation}
    \bar \latent_t(\cdot)=\mathcal \lcg(\latent_t).
\end{equation}
EPCC then uses the play-trained world model to select actions that make maximal progress along this contour while limiting deviation from it.
This follows the central principle of MPCC \cite{krinner_mpcc_2024, lam2010model}.
We formulate this objective using contouring, progress, and action regularization costs.

\textbf{Contouring cost.}
Let $s_{t+k\mid t}\in[0,1]$ denote the progress variable at prediction step $k$.
It selects the corresponding reference state $\bar \latent_t(s_{t+k\mid t})$ along the latent contour.
We define the latent contouring error and its finite-horizon cost as
\begin{align}
    e^c_{t+k\mid t}
    &:=
    \hat \latent_{t+k\mid t}
    -
    \bar \latent_t(s_{t+k\mid t}) \ \regtext{and}
    \label{eq:latent_contouring_error}
    \\
    J_{\mathrm{cont}}(\actionsequence_t,S_t)
    &:=
    \sum_{k=1}^{H}
    \left\|
        e^c_{t+k\mid t}
    \right\|_{Q_c}^{2},\ \regtext{where} \\
    \label{eq:contouring_cost}
    \actionsequence_t
    &:=
    \left(
        \action_{t\mid t},
        \ldots,
        \action_{t+H-1\mid t}
    \right)
\end{align}
is the candidate action sequence and
$S_t=(s_{t\mid t},\ldots,s_{t+H\mid t})$ is the corresponding progress
sequence.


\textbf{Progress cost.}
Let $\nu_{t+k\mid t}\geq0$ denote the progress increment, with
\begin{equation}
    s_{t+k+1\mid t}
    =
    s_{t+k\mid t}
    +
    \nu_{t+k\mid t}.
    \label{eq:progress_dynamics}
\end{equation}
We encourage rapid traversal using
\begin{equation}
    J_{\mathrm{prog}}(V_t)
    :=
    -
    w_p
    \sum_{k=0}^{H-1}
    \nu_{t+k\mid t},
    \label{eq:progress_cost}
\end{equation}
where
$V_t=(\nu_{t\mid t},\ldots,\nu_{t+H-1\mid t})$.
Since each generated contour is anchored at the current state and
$s_{t\mid t}=0$, this is equivalent to
$J_{\mathrm{prog}}=-w_p s_{t+H\mid t}$.

\textbf{Regularization cost.}
We penalize large action magnitudes and abrupt action changes:
\begin{equation}
   J_{\mathrm{act}}(\actionsequence_t)
   :=
   \sum_{k=0}^{H-1}\left\|\action_{t+k\mid t}\right\|_{R_a}^{2}
   +
   \sum_{k=1}^{H-1}\left\|\Delta \action_{t+k\mid t}\right\|_{R_\Delta}^{2},
   \label{eq:action_regularization}
\end{equation}
where
$\Delta \action_{t+k\mid t}
=
\action_{t+k\mid t}-\action_{t+k-1\mid t}$.
The smoothness weight $R_\Delta$ is task-dependent.

\textbf{Optimization.}
At each control step, EPCC jointly optimizes the action sequence and progress evolution:
\begin{equation}
\begin{aligned}
    \actionsequence_t^\star,V_t^\star
    \in
    \underset{\actionsequence_t,V_t}{\arg\min}
    \quad
    &
    J_{\mathrm{cont}}(\actionsequence_t,S_t)
    +
    J_{\mathrm{prog}}(V_t)
    +
    J_{\mathrm{act}}(\actionsequence_t)
    \\[1mm]
    \mathrm{s.t.}
    \quad
    &
    \hat \latent_{t\mid t}
    =
    \encoder(\observation_t),
    \\
    &
    \hat \latent_{t+k+1\mid t}
    =
    \dynamics
    \left(
        \hat \latent_{t+k\mid t},
        \action_{t+k\mid t}
    \right),
    \\
    &
    s_{t+k+1\mid t}
    =
    s_{t+k\mid t}
    +
    \nu_{t+k\mid t},
    \\
    &
    s_{t\mid t}=0,
    \qquad
    0\leq s_{t+k\mid t}\leq1,
    \\
    &
    \action_{t+k\mid t}\in\mathcal A,
    \qquad
    0\leq\nu_{t+k\mid t}\leq\nu_{\max},
    \\
    &
    k=0,\ldots,H-1.
\end{aligned}
\label{eq:epcc_optimization}
\end{equation}
The contouring cost preserves the expert-derived task evolution, while the progress cost drives the predicted rollout forward as quickly as permitted by WM.

After solving Eq.~\eqref{eq:epcc_optimization}, the controller executes the first $T_a$ optimized actions,
\begin{equation}
    \actionsequence_t^{\mathrm{exec}}
    =
    \left(
        \action_{t\mid t}^\star,
        \ldots,
        \action_{t+T_a-1\mid t}^\star
    \right),
    \qquad
    1\leq T_a\leq H,
\end{equation}
and replans from the resulting observation.
At each replanning step, both the latent contour and the optimized action-progress sequence are updated from the latest visual observation.
The full algorithm is in \Cref{alg:expert_play_contouring}.


\begin{algorithm}[t]
\caption{Expert-Play Contouring Control}
\label{alg:expert_play_contouring}
\begin{algorithmic}[1]
\Require Expert demonstrations $\dataset\expert$
\Require Play transitions $\dataset\play$
\Require Planning horizon $H$ and action space $\actionspace$

\Statex \textbf{Offline training}
\State Train encoder $\encoder$ and dynamics $\dynamics$ on
$\dataset\play$ using Eq.~\eqref{eq:wm_total_loss}
\State Freeze $\encoder$ and encode the expert demonstrations
\State Train $\ltg$ on $\dataset\expert$
using Eq.~\eqref{eq:latent_trajectory_generator_loss}
\State Define the latent contour generator
$\mathcal \lcg(z)
\gets
\mathcal P\!\left(z,\ltg(z)\right)$

\Statex \textbf{Online execution}
\State Observe $\observation_0$ and set $t\gets 0$
\While{the task is not terminated}
    \State Encode the current observation:
    $\latent_t\gets \encoder(\observation_t)$
    \State Generate the latent reference contour:
    $\bar \latent_t(\cdot)\gets \lcg(\latent_t)$
    \State Solve \eqref{eq:epcc_optimization} for
    $(\actionsequence_t^\star,V_t^\star)$
    \State Execute the first $T_a$ actions
    \State Observe $\observation_{t+T_a}$ and set $t\gets t+T_a$
\EndWhile
\end{algorithmic}
\end{algorithm}

\section{Simulation Experiments}

\Ours uses play data in two key ways: to model robot-object dynamics under faster execution and to synthesize action beyond temporally rescaled policy outputs.
We therefore study when each of these benefits matters, and how the distribution and amount of play data affects them.
Specifically, we ask:
(1) When does dynamics-aware acceleration help?
(2) When does broader action search from play data help beyond temporal scaling?
(3) How do the distribution and amount of play data affect acceleration?

\subsection{Evaluation Protocol}
\label{sec:exp_protocol}

\textbf{Tasks.}
We evaluate on the \PushT and \Cube tasks from the stable-worldmodel benchmark~\cite{Maes2026stableworldmodelAP}.
\PushT is a planar pushing task that requires the robot to push a T-shaped block to a goal.
\Cube is a pick-and-place task.

We also introduce and evaluate on \LowFricPushT, a controlled variant of \PushT that preserves the task geometry and success condition while varying the sensitivity of object motion to robot interaction through a coasting coefficient $\tau$.
After contact, the block velocity decays as $v(t)=v_0 e^{-t/\tau}$, yielding a stop distance of $v_0\tau$.
Larger $\tau$ therefore increases the effect of each robot-object interaction on subsequent task evolution, making the choice of action increasingly important.
We use $\tau=0$ for quasi-static \PushT setting and $\tau=0.95$ for \LowFricPushT.
This controlled variation lets us isolate how faster-than-demonstration methods behave as robot-object interactions become increasingly consequential.

Per task, we use a set of expert demonstrations and collect play data from several sources, including temporally transformed actions of a policy trained on the demonstrations at a range of faster paces (e.g., down-sampling or faster tracking).
These play trajectories need not complete the task; rather they expose task-relevant robot-object interactions under faster execution. 
\Cref{sec:exp_complementarity} studies which play distribution qualifies for acceleration.

\textbf{Baselines.}
We use \DP \cite{chi2025diffusion} as the imitation baseline.
We compare against three faster-than-demonstration methods:
\SKIP \cite{dai2026skip}, which skips intermediate target poses;
\SAIL \cite{arachchige2025sail}, which shortens the control period outside precision-critical segments; and
\SuP \cite{wu2026speedup}, which uses learned (robot-only) dynamics to select a downsampling rate for action chunks from a frozen policy.
We additionally evaluate \WMFull, which plans with the same play-trained world model but replaces the expert contour with a single goal frame, to test whether play data can solve the task without expert-provided intermediate progression.

\textbf{Metrics.}
We report success rate (SR), time-to-completion over successful episodes (TTC), and throughput (TP) following \SAIL \cite{arachchige2025sail}.
SR measures reliability, TTC measures execution speed conditioned on success, and TP combines both by rewarding faster successes while explicitly penalizing failures.
We define throughput as in \SAIL,
\begin{equation*}
    \mathrm{TP} \;=\; \frac{1}{N}\left(
        \sum_{i \in \mathcal{S}} \frac{1}{\mathrm{TTC}_i}
        \;-\; \frac{N - |\mathcal{S}|}{t_{\max}}
    \right),
\end{equation*}
where $\mathcal{S}$ is the set of successful episodes and $t_{\max}$ is the evaluation time of the task in seconds, the same for every method.
All methods are evaluated on the same $N=50$ initial conditions.

\subsection{When Does Dynamics-Aware Acceleration Help?}
\label{sec:exp_where_play_helps}

\textbf{Hypothesis.}
We hypothesize that \Ours provides the largest advantage when faster execution substantially changes robot-object evolution.
While proxy-based methods such as \SAIL and \SKIP infer when to accelerate from heuristic cues, and \SuP uses learned robot-state dynamics to select a temporal acceleration rate, \Ours directly predicts robot-object evolution under candidate actions.
We therefore expect its gains to be largest on interaction-sensitive tasks such as \LowFricPushT and during contact-rich phases.

\textbf{Setting.}
We first evaluate performances on all manipulation tasks following the protocol in \Cref{sec:exp_protocol}.
To study how the value of dynamics awareness changes with interaction sensitivity, we sweep the coasting time in \LowFricPushT from $\tau=0$ to $4.5$s.
We isolate this effect by keeping the expert-defined task progression fixed: every $\tau$ uses the same quasi-static \PushT demonstrations and only the play data change; the world model and, in its latent space, the contour generator are retrained for each $\tau$.
For phase-level analysis, we use privileged simulator state to separate contact-transition phases, where contact is made or broken, from the remainder of the rollout, and report speedup relative to the demonstration.

\begin{table}[t]
\centering
\caption{\textbf{Main benchmark.}
Metrics are defined in \Cref{sec:exp_protocol}; leading zeros are omitted from TP.
Higher is better for SR and TP, lower for TTC.
TTC is the median over successful episodes: `--' marks a cell with no successful episode.
Bold (resp.\ underline) marks the best (resp.\ second best) entry in each metric column.
}
\label{tab:main_benchmark}
\setlength{\tabcolsep}{1.5pt}
\renewcommand{\arraystretch}{1.0}
\begin{tabular}{@{}l rr>{\columncolor{tphl}}r rr>{\columncolor{tphl}}r rr>{\columncolor{tphl}}r@{}}
\toprule
& \multicolumn{3}{c}{\PushT}
& \multicolumn{3}{c}{\Cube}
& \multicolumn{3}{c}{\LowFricPushT} \\
\cmidrule(lr){2-4}\cmidrule(lr){5-7}\cmidrule(lr){8-10}
& SR & TTC & TP & SR & TTC & TP & SR & TTC & TP \\
& \footnotesize(\%) & \footnotesize(s) & \footnotesize(1/s)
& \footnotesize(\%) & \footnotesize(s) & \footnotesize(1/s)
& \footnotesize(\%) & \footnotesize(s) & \footnotesize(1/s) \\
\midrule
\DP
& 88 & 24.3 & .033
& \underline{96} & 6.8 & .137
& \underline{68} & 28.8 & \underline{.021} \\

\SKIP
& 56 & 20.2 & .019
& 80 & 4.9 & .146
& 24 & 23.6 & -.003 \\

\SAIL{}
& 66 & 19.8 & .033
& \underline{96} & 3.8 & .240
& 46 & \underline{18.5} & .016 \\

\SuP
& \textbf{92} & \textbf{12.1} & \textbf{.070}
& \textbf{100} & \textbf{3.2} & \textbf{.321}
& 48 & 19.8 & .017 \\

\WMFull
& 4 & 7.1 & -.016
& 0 & -- & -.064
& 0 & -- & -.017 \\
\midrule
\Ours
& \underline{90} & \underline{12.2} & \underline{.068}
& \underline{96} & \underline{3.3} & \underline{.282}
& \textbf{74} & \textbf{17.7} & \textbf{.039} \\
\bottomrule
\end{tabular}
\vspace*{-0.4cm}
\end{table}

\begin{figure*}[t]
    \centering
    \begin{subfigure}[t]{0.495\textwidth}
        \centering
        \includegraphics[width=\linewidth]{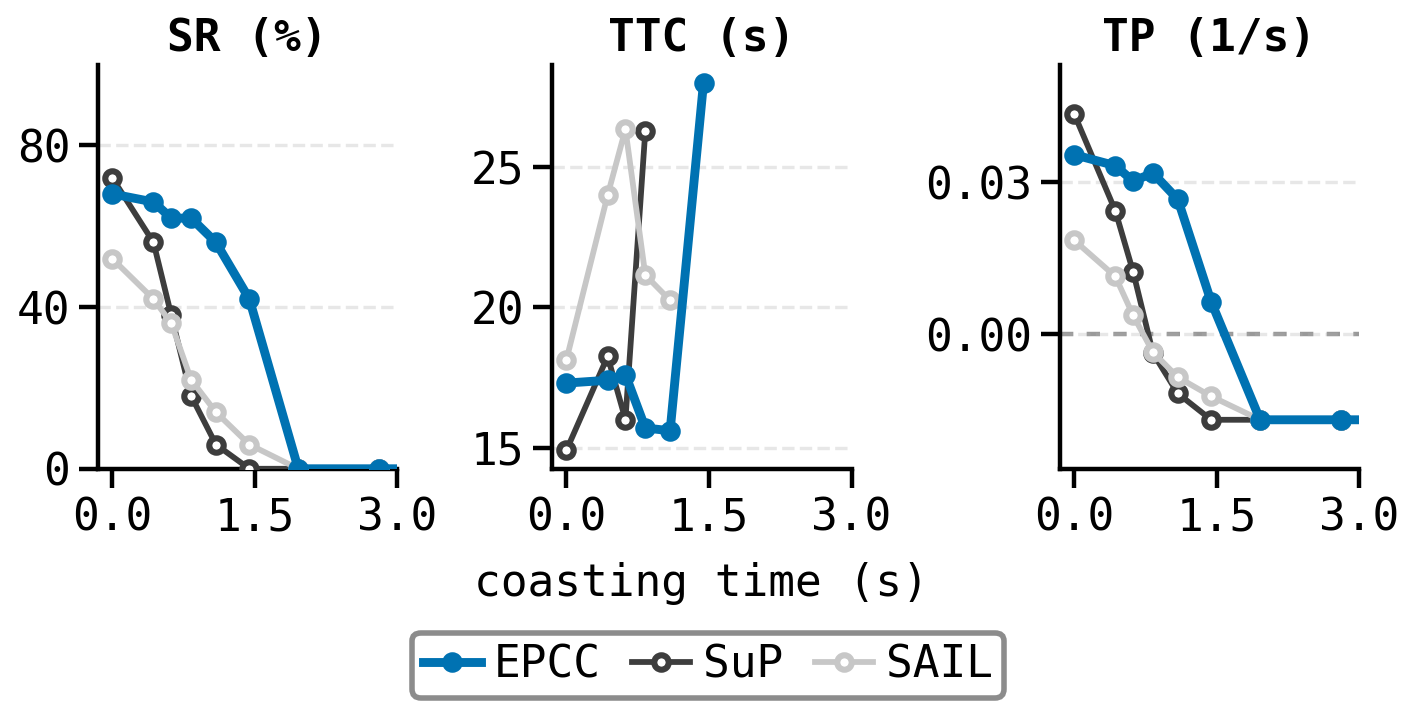}
        \caption{Coasting-time sweep on \LowFricPushT.}
        \label{fig:friction_sweep}
    \end{subfigure}
    \hfill
    \begin{subfigure}[t]{0.495\textwidth}
        \centering
        \includegraphics[width=\linewidth]{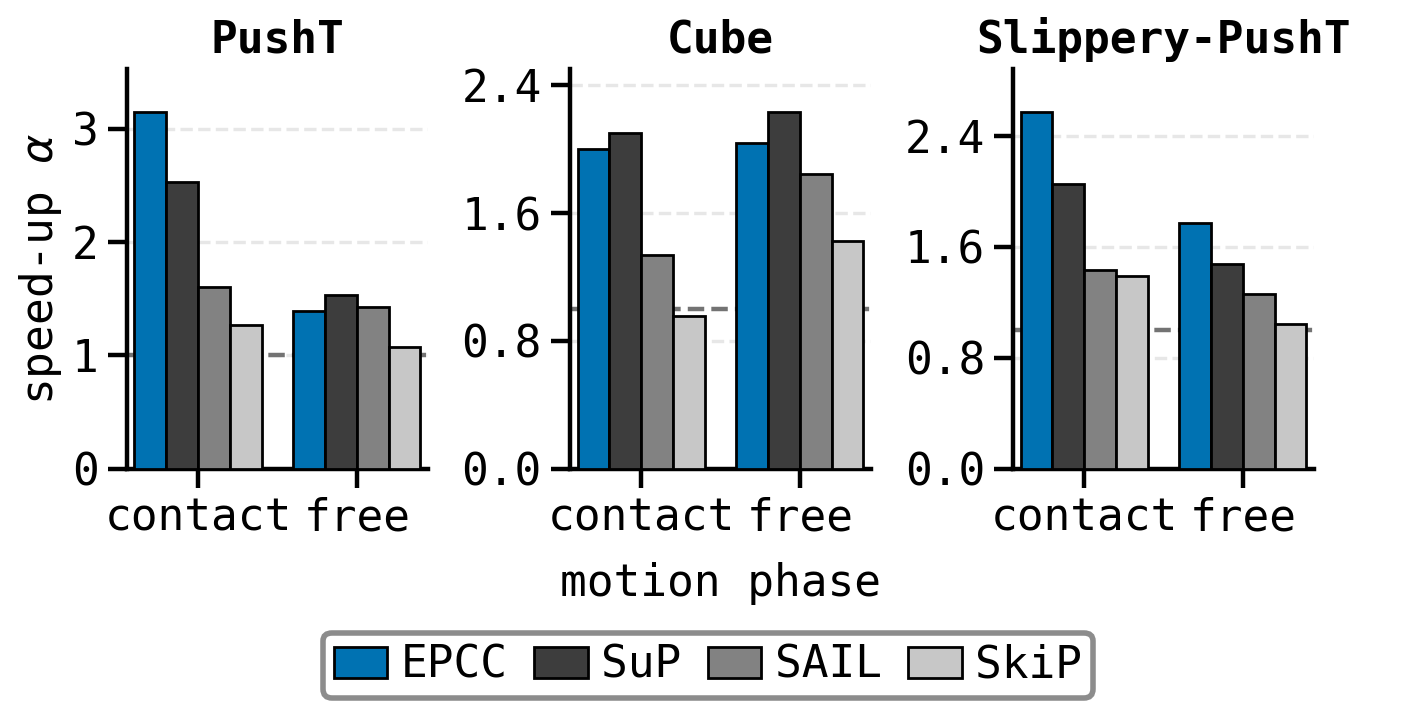}
        \caption{Phase analysis.}
        \label{fig:phase_acceleration}
    \end{subfigure}
    \caption{\textbf{When is dynamics-aware acceleration helpful?}
    (a) SR, time to completion, and TP across the \LowFricPushT with varying block's coasting time $\tau$.
    (b) Speed-up ratio $\alpha$ during contact and free motion, median over each method's successful episodes.
    \Ours gains most as interaction dynamics become more consequential and during contact-sensitive phases.
    }
    \label{fig:dynamics_phase_analysis}
    \vspace*{-0.4cm}
\end{figure*}

\textbf{Results and Discussion.}
At the task level, \Cref{tab:main_benchmark} shows that \Ours provides its clearest gains when faster execution substantially changes robot-object evolution.
Its advantage is largest on the interaction-sensitive \LowFricPushT, where its throughput is $1.9\times$ that of \DP and more than $2\times$ that of every acceleration baseline.
On interaction-sensitive tasks, performance improves with richer dynamics awareness: from proxy-based \SAIL/\SKIP, to robot-state dynamics in \SuP, to robot-object dynamics in \Ours.
 
The controlled sweep in \Cref{fig:friction_sweep} strengthens this trend.
\Ours and \SuP perform similarly in the near-quasi-static regime, but their performance diverges as coasting increases.
At $\tau=1.09$\,s, \Ours still succeeds on $56\%$ of episodes with positive throughput, whereas \SuP and \SAIL fall to $6\%$ and $14\%$ success with negative throughput.
Notably, \Ours achieves these gains without any expert demonstrations from the slippery environments: the same quasi-static \PushT demonstrations are reused throughout the sweep.
That these fixed expert demonstrations remain effective under substantially different interaction dynamics supports our decomposition: expert demonstrations specify task progression, while play provides the dynamics needed to execute it faster.

The phase analysis in \Cref{fig:phase_acceleration} localizes these gains and explains the weaker result on \Cube.
Only $16\%$ of a \Cube episode consists of the interaction-sensitive grasp and release phases;
\Ours matches \SuP there, but is slower over the remaining $84\%$ of the rollout.
Across all three tasks, \Ours averages $2.57\times$ speedup during contact phases, compared with $2.23\times$ for the strongest baseline, while their free-motion speedups are comparable ($1.73\times$ vs.\ $1.75\times$).
Thus, the additional acceleration from dynamics-aware control provided by \Ours is concentrated primarily in phases where faster execution changes robot-object evolution.

\subsection{Does Play Data Provide Useful Actions Beyond Temporal Rescaling?}
\label{sec:exp_select_refine}

Most prior methods accelerate by temporally rescaling actions from a frozen policy.
Here, we ask whether a broader action search provides additional benefit under the same dynamics-aware objective.

\textbf{Hypothesis.}
Temporally rescaled policy actions may suffice when the original motion remains task-valid at higher speed.
When faster execution changes robot-object evolution, however, acceleration may require actions outside this restricted set.
We therefore expect the broader action search enabled by play-trained dynamics to be most beneficial on interaction-sensitive tasks, while temporally rescaled actions may still provide a useful initialization.

\textbf{Setting.}
To isolate the effect of action search, we keep the same dynamics-aware objective and vary only the set of actions the controller can consider.
\textsc{Select} evaluates \SuP-style temporally rescaled action candidates under the \Ours objective and executes the best candidate.
\textsc{EPCC-WS} warm starts optimization from this selected candidate but is free to refine the action sequence beyond the temporal candidate set.
\Ours performs the same optimization without any policy-derived candidates.
Using the same dynamics-aware objective across all variants isolates the benefit of searching beyond temporally rescaled policy actions, as well as whether those actions provide a useful warm start.

\begin{table}[t]
\centering
\caption{\textbf{Play-supported action search vs.\ temporal scaling.}
\textsc{Select} is restricted to temporally rescaled actions from the frozen policy, while \textsc{EPCC-WS} and \Ours can optimize actions beyond this candidate set using the action priors in play data.
Success rate (SR), time-to-completion (TTC), and throughput (TP) are reported over $50$ episodes.}
\label{tab:dof_ladder}
\setlength{\tabcolsep}{1.5pt}
\renewcommand{\arraystretch}{1.0}
\begin{tabular}{@{}l rr>{\columncolor{tphl}}r rr>{\columncolor{tphl}}r rr>{\columncolor{tphl}}r@{}}
\toprule
& \multicolumn{3}{c}{\PushT}
& \multicolumn{3}{c}{\Cube}
& \multicolumn{3}{c}{\LowFricPushT} \\
\cmidrule(lr){2-4}\cmidrule(lr){5-7}\cmidrule(lr){8-10}
& SR & TTC & TP & SR & TTC & TP & SR & TTC & TP \\
& \footnotesize(\%) & \footnotesize(s) & \footnotesize(1/s)
& \footnotesize(\%) & \footnotesize(s) & \footnotesize(1/s)
& \footnotesize(\%) & \footnotesize(s) & \footnotesize(1/s) \\
\midrule
\textsc{Select}
& \underline{92} & 12.6 & .069
& \textbf{96} & 3.5 & .270
& 16 & 24.5 & -.005 \\
\midrule
\textsc{EPCC-WS}
& \textbf{94} & \textbf{10.0} & \textbf{.092}
& 78 & \underline{2.6} & \underline{.281}
& \underline{60} & \underline{20.2} & \underline{.030} \\

\Ours
& 78 & \underline{10.4} & \underline{.073}
& \underline{84} & \textbf{2.5} & \textbf{.317}
& \textbf{62} & \textbf{15.8} & \textbf{.033} \\
\bottomrule
\end{tabular}
\end{table}

\textbf{Results and Discussion.}
Across all tasks, optimizing beyond temporally rescaled policy actions reduces TTC by an average of $28.3\%$, and either \Ours or \textsc{EPCC-WS} achieves the highest throughput.
The benefit is largest on interaction-sensitive \LowFricPushT: \textsc{Select} achieves only $16\%$ success with negative throughput, whereas \textsc{EPCC-WS} and \Ours reach $60\%$ and $62\%$ success with positive throughput.
On \PushT and \Cube, temporally rescaled policy actions are already stronger candidates, but broader action optimization still improves the success-speed trade-off.
This suggests that restricting acceleration to retimed policy outputs becomes increasingly limiting when faster execution changes robot-object evolution.

Temporal candidates can nevertheless provide a useful warm start.
\textsc{EPCC-WS} performs best on \PushT, while \Ours performs better on \Cube and \LowFricPushT, showing that policy-derived candidates can aid initialization but are not required for effective acceleration.

\subsection{What Play Data Is Useful for Acceleration?}
\label{sec:exp_complementarity}

\textbf{Hypothesis.}
We hypothesize that successful acceleration requires complementary information: expert demonstrations define successful task progression, while play captures the robot-object outcomes of faster actions.
We therefore expect that expert data cannot substitute for play in learning World Model, just as play alone cannot replace expert guidance on task progression.
We further expect the value of play to depend on its \emph{pace}, which determines the interaction regimes it covers, and its \emph{amount}, which determines how broadly these outcomes are represented.

\textbf{Setting.}
We examine three aspects of play data. \textit{Complementary roles.} 
We compare \Ours against \textsc{Expert-Only}, which trains the WM on expert rather than play transitions, and \WMFull, which uses play-trained WM but replaces expert contour with a single terminal goal frame.
\textit{Collection Pace.} We train WM on equal-budget play datasets collected at different multiples of demonstration pace.
\textit{Data amount.} We train WM with increasing amounts of play data to assess how \Ours's performance scales with data availability.
For the pace and amount studies, the expert contour generator and the controller remain fixed to isolate the effect of play. 
See Appendix~\ref{app:ctrl:variants} for details.

\begin{figure}[t]
    \centering
    \includegraphics[width=\columnwidth]{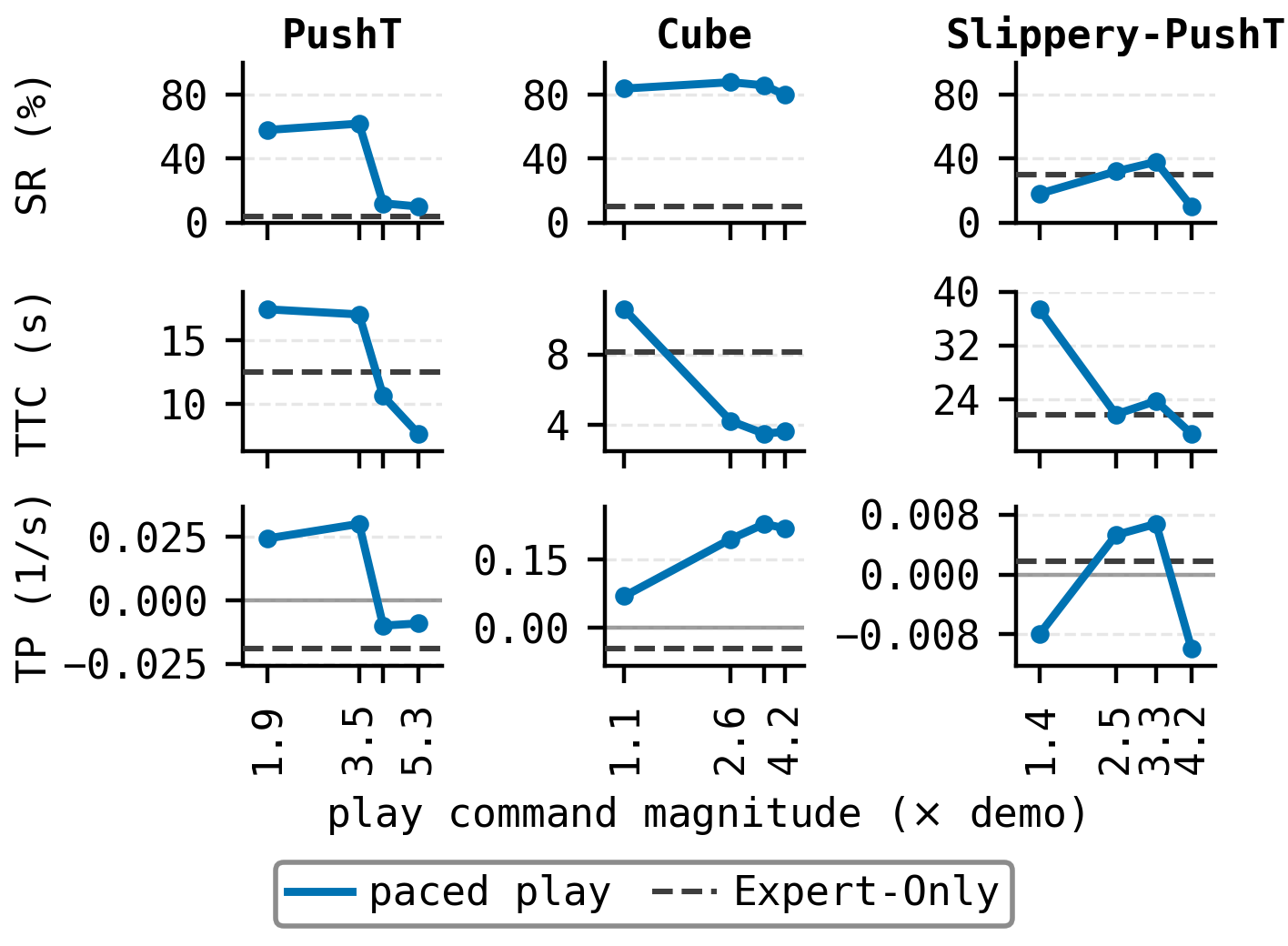}
    \caption{\textbf{What play data enables acceleration?}
    \Ours performance with equal-budget play datasets collected at different paces. Dashed lines denote \textsc{Expert-Only}; Play data at $2.6$-$3.5\times$ demonstration pace achieves highest throughput.}
    \label{fig:play_pace_sweep}
    \vspace*{-0.4cm}
\end{figure}

\textbf{Results and Discussion.}
\textit{Complementary roles.} \textsc{Expert-Only} achieves only $15\%$ average success, compared with $63\%$ using the best play data, and yields negative throughput on both \PushT and \Cube (\Cref{fig:play_pace_sweep}).
Conversely, \WMFull achieves only $1.3\%$ average success without an expert contour (\Cref{tab:main_benchmark}).
Thus, slow demonstrations alone lack coverage of fast interactions, whereas play alone, even with a terminal goal, lacks the guidance for successful task progression.

\textit{Collection pace.} Throughput peaks at intermediate play pace, corresponding to roughly $2.6$-$3.5\times$ the demonstration pace (\Cref{fig:play_pace_sweep}).
At the fastest setting, average success drops from $63\%$ to $33\%$ despite shorter successful completion times, suggesting that overly fast collection can sacrifice task-relevant coverage.

\begin{figure}[t]
    \centering
    \includegraphics[width=\columnwidth]{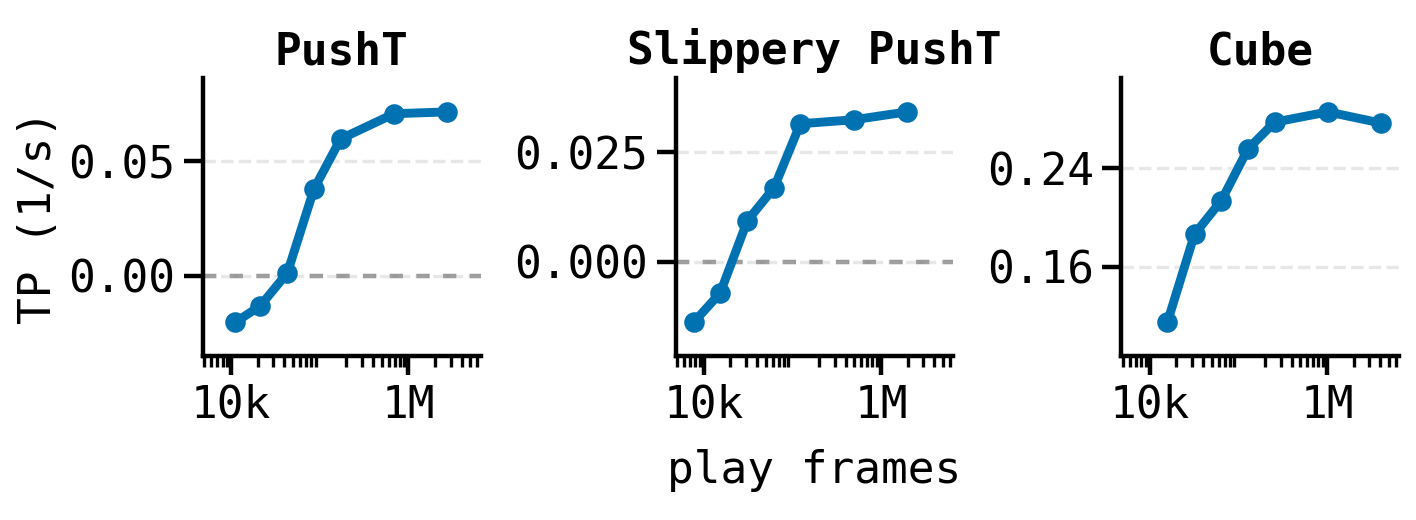}
    \caption{\textbf{Scaling with play data.}
    Throughput of EPCC across tasks as the amount of play data used to train WM increases.}
    \label{fig:play_data_scaling}
\end{figure}

\textit{Data amount.}
Increasing play data improves throughput across all three tasks, with gains plateauing at larger dataset size (\Cref{fig:play_data_scaling}).
These gains primarily reflect more reliable task completion rather than shorter successful episodes.

\section{Real-World Demonstration}
\begin{figure}[htbp]
    \centering
\includegraphics[width=0.8\linewidth]{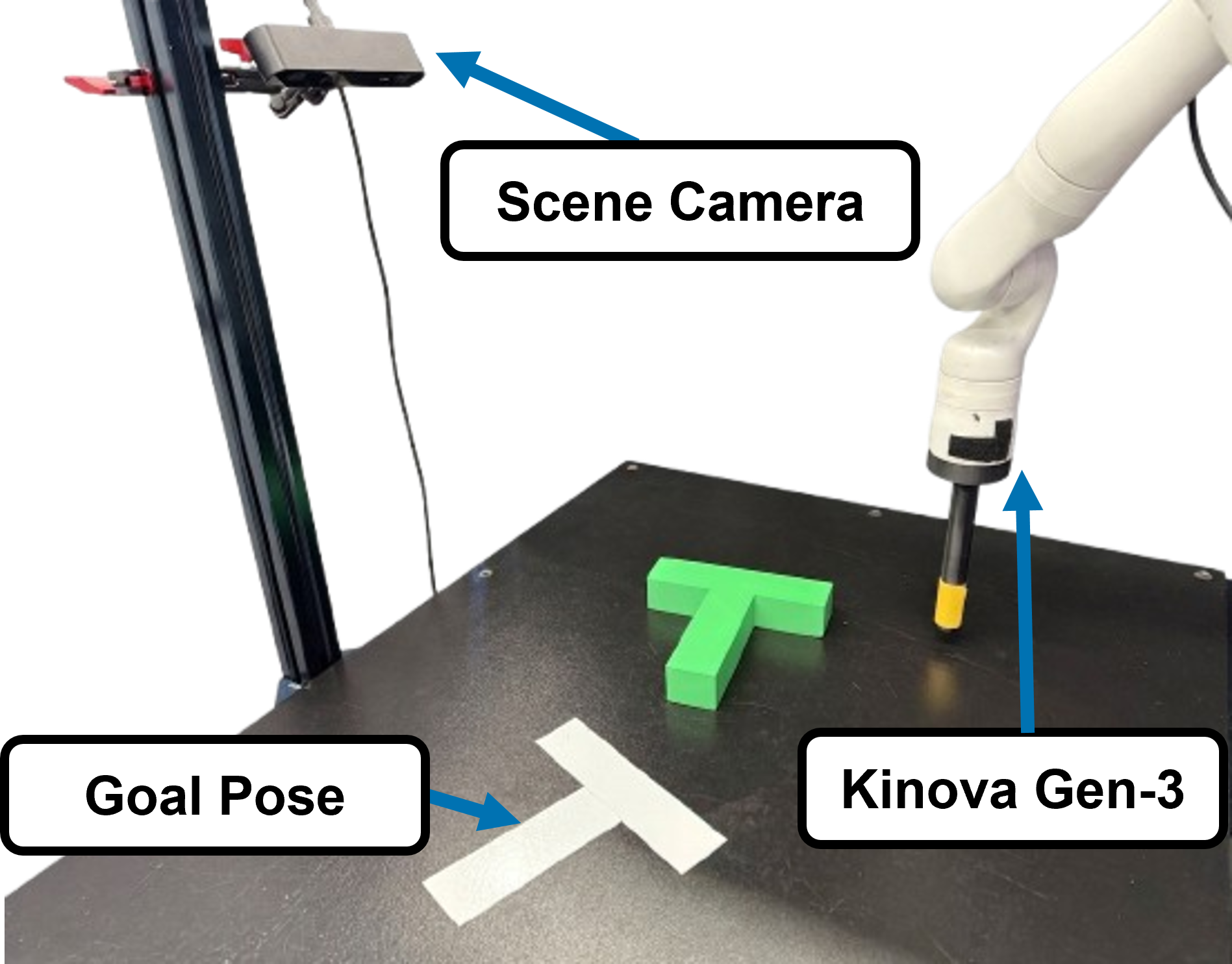}
    \caption{\textbf{Real-world setup.} A Kinova Gen3 arm uses a ZED scene camera to push a T-shaped object into the target region on a low-friction surface, where faster execution makes object motion more sensitive to contact dynamics.}
    \label{fig:real_setup}
\end{figure}

We deploy \Ours on a Kinova Gen3 arm for Push-T on a low-friction table (\Cref{fig:real_setup}), akin to Slippery-PushT: the sliding T-block makes fast teleoperated demonstrations hard to collect, but permits aggressive play.
The robot pushes the block with a rigid rod, observing a fixed camera and its planar end-effector position, and commands planar end-effector displacements at 20\,Hz.
We collect 111 teleoperated
demonstrations (1.2\,h) and 2.7 hours of play data, recorded at $2.2\times$ the demonstrations' median speed. The world model (Sec.~IV-A) is trained on play and demonstrations and the contour generator (Sec.~IV-B) on demonstrations only.
\Cref{fig:teaser} compares a
Flow Matching policy \cite{chi2025diffusion} rollout with \Ours rollout from the same initial state.
In this illustrative example, task success is achieved by \Ours in 13 seconds and by the IL policy in 22 seconds, showing a 1.69x speedup.
Please see our project website for videos and additional demonstrations.
\section{Conclusion}

In this work, we study how fast play can be combined with expert demonstrations for faster-than-demonstration visuomotor manipulation. 
We present Expert-Play Contouring Control (\Ours), which uses expert demonstrations to define successful task progression and fast play to learn the robot-object dynamics encountered beyond demonstration speed. 
To our knowledge, this is the first work to use fast, non-expert play explicitly for dynamics-aware policy acceleration. 
Across three manipulation tasks, \Ours achieves the largest macro-averaged throughput gain over the imitation baseline, with its strongest advantages on interaction-sensitive tasks.
Notably, we show \Ours scales with more play data.

\textbf{Limitations.}
\Ours depends on the accuracy and coverage of its world model and incurs additional planning cost. 
More targeted play-data collection and faster world models may improve both reliability and real-time performance.
Jointly generating contours and actions is another promising direction \cite{ye2026world, jung2025joint}.



\section*{ACKNOWLEDGMENT}
We thank Long Kiu Chung for helpful discussions.

\bibliography{references}  

\clearpage
\appendices

\section{EPCC Implementation Details}

In this section, we provide the implementation details of Latent World Model, Latent Contour Generator, and Latent Contouring Controller.

\label{sec:app:implementation}

\subsection{Latent World Model}
\label{sec:app:training_wm}
We adopt LeWM~\cite{maes2026leworldmodel} with a ViT-Tiny
visual encoder, $224\times224$ image inputs, a context length of three frames, and an action block size of five.
\PushT and \LowFricPushT use a single camera, while
\Cube uses scene and wrist cameras.
For \Cube, we weight the scene and wrist prediction
losses by $1.0$ and $1.5$, respectively.
We train the world models using AdamW with a learning rate of \(5\times10^{-5}\), for 15 epochs on the pushing tasks and 7 epochs on Cube.

\subsection{Expert Latent Contour Generator}
\label{sec:app:generator}

We encode expert demonstrations using the learned LeWM visual encoder and projector to obtain the latent demonstrations.
We train a conditional 1D U-Net diffusion model implemented in \textsc{CleanDiffuser}~\cite{dong2024cleandiffuser} to model $p_\psi(z_{t+1:t+H-1}\mid z_{t-K+1:t})$, with an observation horizon $K=6$ and a prediction horizon of $H=84$.

We use clean-sample ($x_0$) prediction rather than noise prediction, which was less stable in our experiments, and add temporal-displacement supervision:
\begin{equation}
\begin{aligned}
\mathcal L_{\mathrm{LTG}} = \frac{1}{H-1}
\sum_{k=1}^{H-1}
\Big[&\|\hat x_0^k-x_0^k\|_2^2\\
&+\lambda_{\mathrm{vel}}
\|\Delta\hat x_0^k-\Delta x_0^k\|_2^2\Big],
\end{aligned}
\label{eq:app:generator_training_loss}
\end{equation}
where $\Delta x_0^k=x_0^k-x_0^{k-1}$, $\Delta\hat x_0^k=\hat x_0^k-\hat x_0^{k-1}$, and $\lambda_{\mathrm{vel}}=3$.
We apply per-dimension min-max normalization and use a 50-step DDPM with a cosine noise schedule.
We train using AdamW with a learning rate of $2\times10^{-4}$.

\subsection{Latent Contouring Controller}
\label{sec:app:controller}

Given the trained world model and expert contour generator,
we solve Eq.~\eqref{eq:epcc_optimization} using the
cross-entropy method (CEM).
We use a planning horizon of five action blocks, execution horizon of three on \PushT{} and \Cube{} or one block on \LowFricPushT{}, where one action block corresponds to five actions.

We parameterize progress relative to the reference pace:
$\nu_k=(5/84)\tilde\nu_k$, where $\tilde\nu_k=1$
corresponds to the nominal pace of the 84-frame expert
reference. The implemented objective is
\begin{equation}
\begin{aligned}
J ={}&
q_c\sum_{k=1}^{H}
\|\hat z_k-\bar z_t(s_k)\|_2^2
-w_p\sum_{k=0}^{H-1}\tilde\nu_k \\
&+q_a\sum_{j=1}^{5H}\|\bar a_j\|_2^2
+r_\Delta\sum_{j=2}^{5H}
\|\bar a_j-\bar a_{j-1}\|_2^2,
\end{aligned}
\label{eq:app:controller_implemented}
\end{equation}
where $H=5$ and $\bar a_j$ denotes a normalized
primitive action.
We use $q_c=0.107$ and $q_a=5.8\times10^{-3}$
across tasks.
The smoothness term penalizes adjacent actions within
the planned sequence and is disabled on \PushT{}.
CEM uses 30 iterations and 30 elites, with action plans
warm-started from the previous solution.
Task-specific settings are in Table~\ref{tab:ctrl-hparams}.

\begin{table}[H]
\centering\footnotesize
\setlength{\tabcolsep}{4pt}
\caption{Main-benchmark controller settings.}
\label{tab:ctrl-hparams}
\begin{tabular}{lccc}
\toprule
 & \PushT & \LowFricPushT & \Cube \\
\midrule
Execution Horizon & $3$ & $1$ & $3$ \\
CEM candidates & $300$ & $300$ & $1000$ \\
Progress weight $w_p$ & $4.21$ & $9.26$ & $6.01$ \\
Progress cap $\tilde\nu_{\max}$ & $4.0$ & $2.7$ & $4.0$ \\
Smoothness $r_\Delta$ & $0$ & $1$ & $1$ \\
\bottomrule
\end{tabular}
\end{table}

\section{Additional Experimental Analysis}
\label{sec:app:analysis}

\subsection{Phase-Specific Speed-Up Analysis (\Cref{fig:phase_acceleration})}
\label{sec:app:phase_definitions}
We provide the phase definitions and evaluation protocol for the analysis in \Cref{sec:exp_where_play_helps} and \Cref{fig:phase_acceleration}.

\textbf{Phase Definitions.}
We use privileged simulator states only to partition evaluation rollouts into contact-sensitive and free-motion intervals.
For \PushT{} and \LowFricPushT{}, contact-sensitive phase is when the distance from the pusher center to the block
surface is within the pusher radius (15 pixels).
For \Cube{}, we use the expert oracle's event thresholds to divide each rollout into \textsc{above}, \textsc{descend}, \textsc{grasp}, \textsc{lift}, \textsc{carry}, and \textsc{place} phase groups, with \textsc{grasp} and \textsc{place} labeled as contact-sensitive.

\textbf{Evaluation Protocol.}
For each phase group $g$, we compute the speed-up
$\alpha_{i,g}=T_{i,g}^{\mathrm{demo}}/
T_{i,g}^{\mathrm{method}}$ for each initial condition $i$
and report the median over initial conditions where each
method (\Ours{}, \textsc{SkiP}, \SAIL{}, and \SuP{})
succeeds.
This yields 45/28/33/46, 45/38/46/45, and 37/12/23/24 initial
conditions for \PushT{}, \Cube{}, and \LowFricPushT{},
respectively.

\subsection{Friction Sweep (\Cref{fig:friction_sweep})}
\label{sec:app:friction_sweep}
For the friction sweep in \Cref{sec:exp_where_play_helps}, we vary block damping over coasting time $\tau\in[0,4.5]$\,s.
At each setting, we train a separate world model, including its visual encoder, on play data from that environment.
We use the expert contour generator using the same \PushT{} demonstrations encoded in the corresponding latent space.
\Ours{} uses the \LowFricPushT{} controller with $\tilde\nu_{\max}=2.2$ and $r_\Delta=3$.
\SAIL{} uses $k=2$; 
the \SAIL{} and \SuP{} policies are not retrained across damping settings.

\subsection{Action-Search Comparison (\Cref{tab:dof_ladder})}
For the comparison in \Cref{sec:exp_select_refine}, \Ours{}, \textsc{Select}, and \textsc{EPCC-WS} use a planning horizon of three action blocks, an execution horizon of one block, progress cap $\tilde\nu_{\max}=16$.
\textsc{Select} and \textsc{EPCC-WS} generate candidates by rescaling the first 75 actions of a 128-action diffusion-policy chunk using $k\in\{1,\ldots,5\}$, rather than the 15-action chunks used by \SuP{}.
\textsc{Select} executes the lowest-cost candidate under the EPCC objective; 
\textsc{EPCC-WS} uses that candidate to initialize CEM.

\subsection{Play Collection Pace (\Cref{fig:play_pace_sweep})}
\label{sec:app:play_pace}
For the pace study in \Cref{sec:exp_complementarity}, we measure collection pace as the 90th-percentile action magnitude relative to expert demonstrations.
The \Cube{} controller uses $\tilde\nu_{\max}=2.8$ at the nominal setting;
\LowFricPushT{} uses $\tilde\nu_{\max}=2.2$ and $r_\Delta=3$.
Across pace settings, the controller's progress cap, initial progress mean, and progress standard deviation change with the collection setting, while the progress reward weight is scaled inversely.

\subsection{Play-Data Scaling Analysis (\Cref{fig:play_data_scaling})}
\label{app:play_data_scaling}

\begin{figure}[t]
    \centering
    \includegraphics[width=\columnwidth]{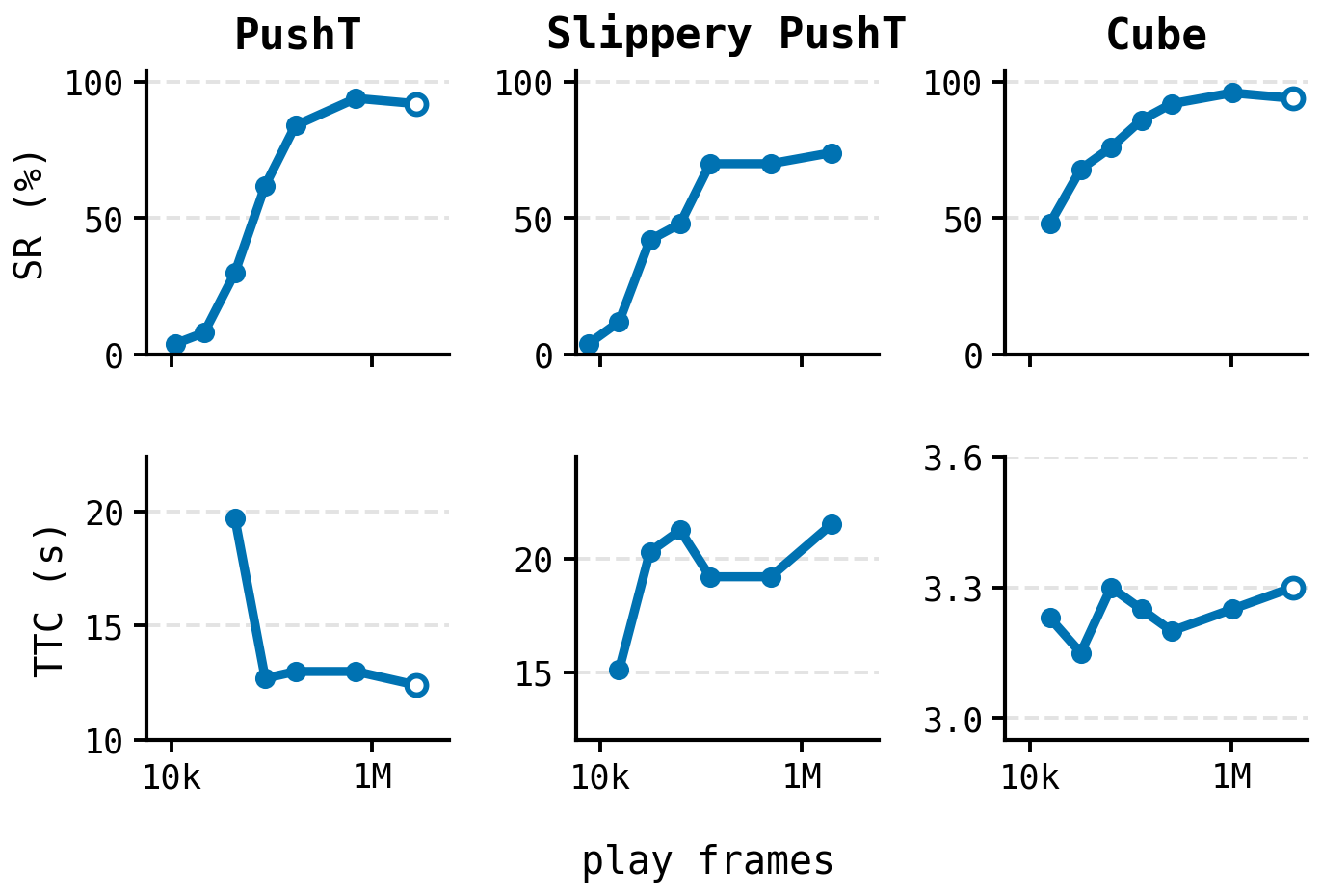}
    \caption{\textbf{Additional play-data scaling results.}
    Success rate (top) and median completion time among successes (bottom) across dataset sizes; the corresponding throughput results appear in \Cref{fig:play_data_scaling}.
    TTC is omitted for fewer than five successful episodes.}
    \label{fig:app:scaling_sr_ttc}
\end{figure}

We provide the training and evaluation details for the play-data scaling experiment in \Cref{sec:exp_complementarity}, with additional
success-rate and completion-time results in \Cref{fig:app:scaling_sr_ttc}.

\textbf{Data and Training.}
We use the aggregated play datasets for each task ($O+B+D$), containing $2.76$, $1.97$, and $4.10$ million frames for \PushT{}, \LowFricPushT{}, and \Cube{}, respectively.
We randomly sample seven subsets with fractions $2^{-k}$, $k\in\{0,2,4,5,6,7,8\}$, preserving the mixture of play sources and collection paces.

For each task, we reuse the pretrained LeWM encoder $\encoder$ and the expert trajectory generator $\ltg$ used in the main benchmark (\Cref{tab:main_benchmark}).
We keep both fixed and train a new dynamics predictor $\dynamics$ from scratch on each subset.
This isolates the effect of play-data amount on dynamics learning, independent of changes to the latent representation or expert trajectory generator.

We select each predictor using validation prediction loss and evaluate it on the same $50$ initial conditions.
\Cref{fig:play_data_scaling} reports throughput (TP), while \Cref{fig:app:scaling_sr_ttc} additionally reports success rate (SR) and median time to completion (TTC) among successful episodes.

\textbf{Results.}
As shown in \Cref{fig:app:scaling_sr_ttc}, increasing play data consistently improves success rates, while completion times show no consistent reduction.
This suggests that the throughput gains in \Cref{fig:play_data_scaling} primarily arise from more reliable task completion rather than faster successful executions.
These results are consistent with broader coverage of robot-object dynamics improving the reliability of faster-than-demonstration execution.

\subsection{Controlled-Study Settings}
\label{app:base:variants}
\textbf{Friction sweep.}
For \Cref{fig:friction_sweep}, we train a separate world model,
including its encoder, at each damping setting
($d\in[0,0.8]$, equivalent to $\tau\in[0,4.5]$\,s).
The expert contour generator is retrained using the same
\PushT{} demonstrations encoded in the corresponding latent space.
EPCC uses the \LowFricPushT{} controller with
$\tilde\nu_{\max}=2.2$ and $r_\Delta=3$.
\SAIL{} uses $k=2$ instead of its main-benchmark value of $2.5$;
\SuP{} and \SAIL{} policies are not retrained across friction levels.
For the phase analysis (\Cref{fig:phase_acceleration}),
\SAIL{} uses its main-benchmark values of $k$.

\textbf{Action-search comparison.}\label{app:ctrl:variants}
For \Cref{tab:dof_ladder}, EPCC, \textsc{Select}, and
\textsc{EPCC-WS} share a planning horizon of $H=3$ action
blocks, an execution horizon of one block, progress cap
$\tilde\nu_{\max}=16$, action bounds $[-2,2]$, and the
$O+B+D$ \LowFricPushT{} world model.
\textsc{Select} and \textsc{EPCC-WS} rescale the first
$75$ actions of a $128$-action diffusion-policy chunk using
$k\in\{1,\ldots,5\}$; unlike SuP, these do not use the
15-action chunks described in Appendix~\ref{app:base:accel}.
\textsc{Select} picks the lowest-cost candidate under EPCC's
objective; \textsc{EPCC-WS} uses that candidate to warm-start CEM.

\textbf{Play collection pace.}
For \Cref{fig:play_pace_sweep}, the \Cube{} controller uses
action bounds $[-1,1]$ and $\tilde\nu_{\max}=2.8$;
\LowFricPushT{} uses $\tilde\nu_{\max}=2.2$ and
$r_\Delta=3$.
The measured collection pace is the $90$th-percentile action
magnitude relative to the expert demonstrations.
Controller progress cap, initial progress mean, and progress
standard deviation change with the collector's pace, while
the progress reward weight is divided by it.
\Cref{tab:ctrl-fig4} gives the measured pace and progress cap.
\begin{table}[H]
\centering\footnotesize
\setlength{\tabcolsep}{4pt}
\caption{Play collection pace (relative to expert demonstrations)
and corresponding controller progress cap for
\Cref{fig:play_pace_sweep}. Entries are pace / $\tilde\nu_{\max}$;
$^\dagger$ denotes the nominal $1\times$ play condition.}
\label{tab:ctrl-fig4}
\begin{tabular}{lccc}
\toprule
& \PushT{} & \LowFricPushT{} & \Cube{} \\
\midrule
Expert-Only & --- / $4.0$ & --- / $2.2$ & --- / $2.8$ \\
Play 1 & $1.85$ / $2.0$ & $1.44$ / $1.1$ & $1.06$ / $0.84$ \\
Play 2 & $3.51^\dagger$ / $4.0$ & $2.47^\dagger$ / $2.2$ & $2.65$ / $2.1$ \\
Play 3 & $4.15$ / $6.0$ & $3.28$ / $3.3$ & $3.54^\dagger$ / $2.8$ \\
Play 4 & $5.30$ / $10.0$ & $4.21$ / $5.5$ & $4.22$ / $3.4$ \\
\bottomrule
\end{tabular}
\end{table}

\FloatBarrier

\section{Simulation Tasks and Evaluation Details}
\label{sec:app:exp_details}

\subsection{Simulation Tasks and Success Criteria}
\label{sec:app:tasks}
\begin{figure}[t]
    \centering
    \begin{subfigure}[t]{0.47\columnwidth}
        \centering
        \includegraphics[width=\linewidth]{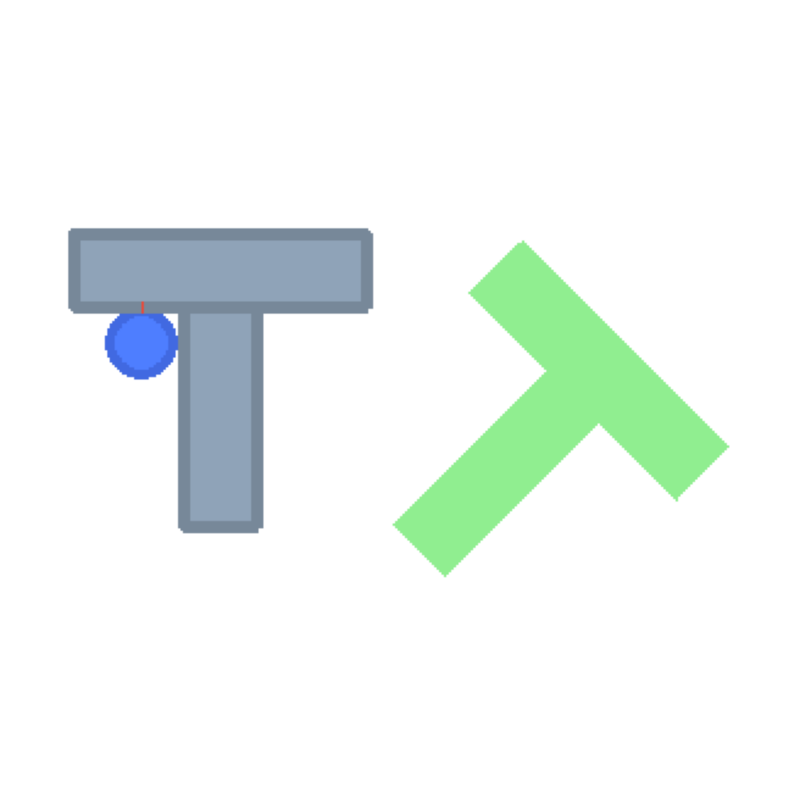}
        \caption{\PushT{} / \LowFricPushT{}}
        \label{fig:app:task_pusht}
    \end{subfigure}\hfill
    \begin{subfigure}[t]{0.47\columnwidth}
        \centering
        \includegraphics[width=\linewidth]{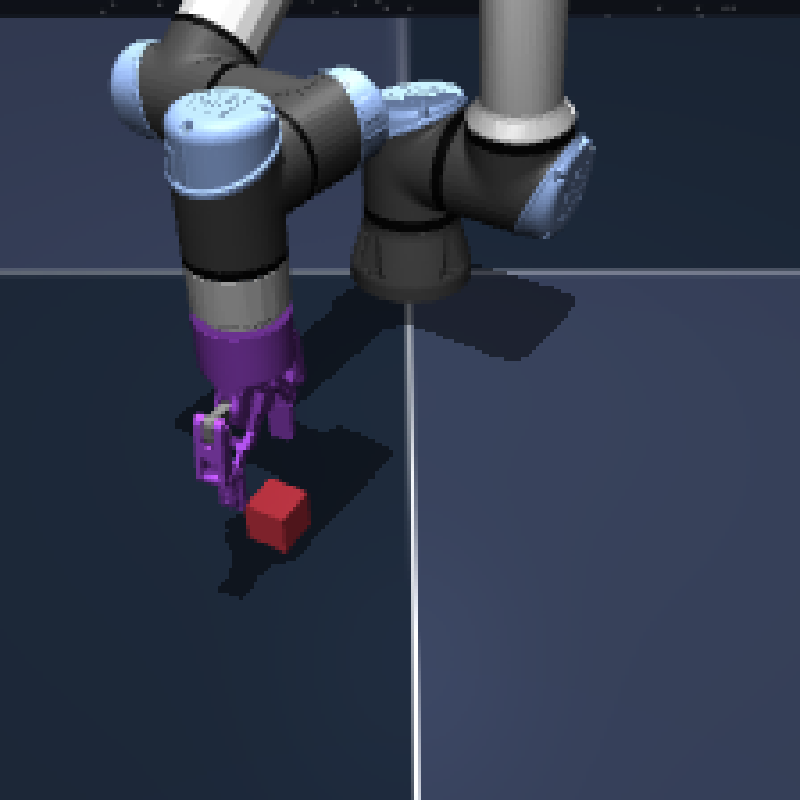}
        \caption{\Cube{}}
        \label{fig:app:task_cube}
    \end{subfigure}
    \caption{\textbf{Simulation tasks.} \PushT{} and \LowFricPushT{}
    share the same pushing geometry but differ in block damping;
    \Cube{} requires grasping and placing an object.}
    \label{fig:app:tasks}
\end{figure}

\textbf{\PushT{}}.
We adopt the planar pushing environment used in
LeWM~\cite{maes2026leworldmodel}, where a circular pusher with $2D$ action space moves a T-shaped block to a goal pose in a $512\times512$ \texttt{pymunk} workspace. 
The pusher has a $15$-pixel radius and receives planar relative-position actions at $10$\,Hz. 
We remove containment walls so that the block can overshoot the goal. 
Success requires the block to remain within $20$ pixels and $\pi/9$ radians of the goal pose for five consecutive seconds.

\textbf{\LowFricPushT{}}.
We modify \PushT{} by reducing block damping while retaining the same geometry, actions, and success criterion. 
At each simulation substep of duration $\Delta t$, the block velocity is multiplied by $d^{\Delta t}$, yielding
\begin{equation}
    v(t)=v_0 d^t=v_0e^{-t/\tau},\qquad
    \tau=\frac{1}{\ln(1/d)}.
    \label{eq:app:coasting_time}
\end{equation}
Here $d$ is the velocity fraction retained per second and $\tau$ is the coasting time constant. Standard \PushT{} uses $d=0$ ($\tau=0$ in the limiting case), whereas \LowFricPushT{} uses $d=0.35$ ($\tau\approx0.95$\,s). After contact ends, the block travels an additional distance $v_0\tau$, making fast execution more sensitive to contact velocity.

\textbf{\Cube{}}.
We adopt the pick-and-place task from OGBench~\cite{park2025ogbench,todorov2012mujoco}.
The robot uses five-dimensional end-effector and gripper actions at $20$\,Hz, with scene and wrist cameras. Success requires grasping the cube (raised at least $2$\,cm above the table and within $5$\,cm of the end effector for three consecutive steps), avoiding a subsequent drop, and finishing within $7$\,cm of the goal.

\subsection{Expert Demonstrations and Play Data}
\label{sec:app:dataset_details}

We collect expert demonstrations for training the contour
generator and three sources of play data for training the
world model: uncurated play ($O$), rollouts of \SuP{} at
fixed acceleration factors ($B$), and additional play
collected at different execution paces ($D$).

\textbf{\PushT{}}.
We collect $200$ successful expert demonstrations using
the pretrained Diffusion Policy released by
LeRobot~\cite{cadene2024lerobot}
(\href{https://huggingface.co/lerobot/diffusion_pusht}{checkpoint}).
For uncurated play ($O$), we use the
\href{https://huggingface.co/datasets/quentinll/lewm-pusht}
{PushT dataset} released with LeWM~\cite{maes2026leworldmodel}.
Additional paced play ($D$) is collected by skipping
policy action targets.

\textbf{\LowFricPushT{}.}
We collect $200$ successful expert demonstrations
using a Diffusion Policy that we train in the slippery environment.
For uncurated play ($O$), we roll out a Diffusion Policy
trained on standard \PushT{} in the slippery environment,
retaining both successful and unsuccessful episodes.
We additionally collect paced play ($D$) through
policy-target skipping.

\textbf{\Cube{}.}
We collect $800$ successful expert demonstrations
using a scripted expert motion planner.
For uncurated play ($O$), we execute the scripted
oracle toward randomly sampled goals.
Additional paced play ($D$) is collected by scaling
the oracle's action commands.

For the pushing tasks, we remove wall-contact segments
from $O$ and $D$, while $B$ is collected without walls.
The main benchmark uses $O+B+D$ for \PushT{} and \Cube{},
and $O+B$ for \LowFricPushT{}
(\Cref{tab:app:corpora}).

\begin{table}[t]
\centering\footnotesize
\setlength{\tabcolsep}{3pt}
\caption{Main-benchmark play mixtures (hours) used in
\Cref{tab:main_benchmark}.}
\label{tab:app:corpora}
\begin{tabular}{lccccc}
\toprule
Task & $O$ & $B$ & $D$ & Total & Episodes \\
\midrule
\PushT{} & $62.7$ & $8.0$ & $6.9$ & $77.6$ & $21{,}393$ \\
\LowFricPushT{} & $30.5$ & $16.4$ & --- & $46.8$ & $7{,}880$ \\
\Cube{} & $27.9$ & $18.9$ & $10.7$ & $57.5$ & $19{,}111$ \\
\bottomrule
\end{tabular}
\end{table}

\subsection{Evaluation Protocol}
\label{sec:app:metrics}
\textbf{Initial Conditions and Episode Budget.}
For each task, all methods are evaluated on the same $50$ initial conditions. 
Rollouts continue for their full episode budgets even after satisfying a success criterion. 
The nominal budgets are $435$ steps ($43.5$\,s) for \PushT{}, $594$ steps ($59.4$\,s) for \LowFricPushT{}, and $314$ steps ($15.7$\,s) for \Cube{}. 

\textbf{Evaluation Metrics.}
We report success rate (SR), median time to completion (TTC) among successful episodes, and failure-penalized throughput (TP), following \Cref{sec:exp_protocol} and \cite{arachchige2025sail}:
\begin{equation}
\mathrm{TP}=\frac{1}{N}\left(
\sum_{i\in\mathcal S}\frac{1}{\mathrm{TTC}_i}
-\frac{N-|\mathcal S|}{t_{\max}}\right),
\end{equation}
where $\mathcal S$ denotes successful episodes and $t_{\max}$ is the task's episode budget in seconds.

\FloatBarrier
\section{Baseline Implementation Details}
\label{sec:app:baselines}

\subsection{Diffusion-Policy}
\label{app:base:dp}
We use a ResNet-18 visual encoder per camera, two $96\times96$ images per camera with $84\times84$ crop, and a conditional 1D U-Net with $50$ DDPM steps. 
The policy predicts $16$-action chunks and executes $15$ actions before re-planning.

\subsection{Acceleration Baselines}
\label{app:base:accel}
\textbf{SkiP.}
We select demonstration key poses using high-frequency DCT content
(16-frame window, $0.75$ quantile), turns, stops, and \Cube{}
gripper transitions. The policy is retrained to predict absolute
key poses, which a feedback controller converts to bounded relative
actions in place of SkiP's task-specific reaching controller.

\textbf{SAIL.}
We train the policy to predict reached poses and precision indicators
obtained from AWE waypoints and DBSCAN clustering. Outside precise
segments, the simulation control period is divided by a speed factor
$k\in\{1.5,2,2.5,3\}$, selected on common evaluation conditions
($k=2.5/2/3$ for \PushT{}/\LowFricPushT{}/\Cube{}).
Our adaptation uses the simulator's low-level controller and does
not include SAIL's error-adaptive diffusion guidance, hardware
tracking controller, or latency handling.

\textbf{SuP.}
SuP sums groups of $k\in\{2,3,4\}$ actions from a $15$-action DP
chunk to create accelerated candidates. A dynamics model labels
candidates using maximum interpolated trajectory discrepancy
(40th-percentile threshold for $k>1$), and an IQL scheduler
($\Omega=5$, $\gamma=0.9$, expectile $0.95$) selects the factor.
The scheduler and safety labels use only the robot's simulator
state; the action policy remains image-conditioned.

\subsection{Play+Goal}
\label{app:base:playgoal}
We replace EPCC's expert contour with the terminal goal latent
$z_{\mathrm{goal}}$ from the matched expert demonstration while
keeping the play-trained world model, CEM planner, planning and
execution horizons, and action limits fixed. The planner minimizes
\begin{equation*}
\|\hat z_H-z_{\mathrm{goal}}\|_2^2
+0.01\sum_j\|\bar a_j\|_2^2
+\sum_j\|\bar a_j-\bar a_{j-1}\|_2^2.
\end{equation*}
The goal latent corresponds to the first sustained goal state.



\addtolength{\textheight}{-12cm}   

\end{document}